\documentclass[letterpaper, 10 pt, conference]{ieeeconf}  

\usepackage{graphicx}
\usepackage{lettrine}
\usepackage{amsmath}
\usepackage{booktabs}
\usepackage{multirow}
\usepackage{makecell}
\usepackage{caption}

\IEEEoverridecommandlockouts                              

\title{\LARGE \bf
RadarNeXt: Real-Time and Reliable 3D Object Detector Based On 4D mmWave Imaging Radar
}

\author{Liye Jia$^{1, 2,3, 4,\dagger}$, Runwei Guan$^{6,\dagger}$, Haocheng Zhao$^{1, 2,3, 4,\dagger}$, Qiuchi Zhao$^{5}$, Ka Lok Man$^{2}$, \\ Jeremy Smith$^{3}$, Limin Yu$^{2}$, Yutao Yue$^{7, 1, 4, *}$
\thanks{$^{1}$Institute of Deep Perception Technology, JITRI, Wuxi, China}
\thanks{$^{2}$SAT, Xi’an Jiaotong-Liverpool University, Suzhou, China}
\thanks{$^{3}$Department of EEE, University of Liverpool, Liverpool}
\thanks{$^{4}$XJTLU-JITRI Academy of Industrial Technology, Xi’an Jiaotong-Liverpool University, Suzhou, China}
\thanks{$^{5}$School of Automation Science and Electrical Engineering, Beihang University, Beijing, China} 
\thanks{$^{6}$Thrust of Artificial Intelligence, HKUST (GZ), Guangzhou, China}
\thanks{$^{7}$Thrust of Artificial Intelligence and Thrust of Intelligent Transportation, HKUST (GZ), Guangzhou, China.}
\thanks{$^\dagger$Liye Jia, Runwei Guan, and Haocheng Zhao contribute equally.}
\thanks{$^{*}$Corresponding author: yutaoyue@hkust-gz.edu.cn}
}

\begin{document}

\maketitle
\thispagestyle{empty}
\pagestyle{empty}

\begin{abstract}

3D object detection is crucial for Autonomous Driving (AD) and Advanced Driver Assistance Systems (ADAS). However, most 3D detectors prioritize detection accuracy, often overlooking network inference speed in practical applications. In this paper, we propose RadarNeXt, a real-time and reliable 3D object detector based on the 4D mmWave radar point clouds. It leverages the re-parameterizable neural networks to catch multi-scale features, reduce memory cost and accelerate the inference. Moreover, to highlight the irregular foreground features of radar point clouds and suppress background clutter, we propose a Multi-path Deformable Foreground Enhancement Network (MDFEN), ensuring detection accuracy while minimizing the sacrifice of speed and excessive number of parameters. Experimental results on View-of-Delft and TJ4DRadSet datasets validate the exceptional performance and efficiency of RadarNeXt, achieving 50.48 and 32.30 mAPs with the variant using our proposed MDFEN. Notably, our RadarNeXt variants achieve inference speeds of over 67.10 FPS on the RTX A4000 GPU and 28.40 FPS on the Jetson AGX Orin. This research demonstrates that RadarNeXt brings a novel and effective paradigm for 3D perception based on 4D mmWave radar.

\end{abstract}


\section{INTRODUCTION}

\lettrine{O}{bject} detection is fundamental to Autonomous Driving (AD) and Advanced Driver-Assistance Systems (ADAS) for scene perceptions \cite{ma20233d}. However, traditional 2D object detection, limited by cameras, lacks accurate 3D information for downstream tasks \cite{wang2024comprehensive}. Therefore, point clouds from LiDAR or millimeter-wave (mmWave) radar are crucial for 3D detection in perception systems \cite{yao2023radarcam}. As an emerging 3D sensor, 4D mmWave radar provides valuable information, including real 3D coordinates, Radar Cross Section (RCS), and relative velocity \cite{yao2023radar}. Moreover, compared to LiDAR, radar has a longer detection range and is more robust to driving conditions, such as adverse weather \cite{yao2024waterscenes}. The similarity between 4D mmWave radar and LiDAR point clouds allows for directly applying existing LiDAR-based networks to radar-based 3D object detection \cite{jiang20234d}. However, the extreme sparsity and random noise in 4D mmWave radar point clouds hinder the performance of these networks \cite{venon2022millimeter}.

\begin{figure}[!t]
\centering
\includegraphics[width=3.2in]{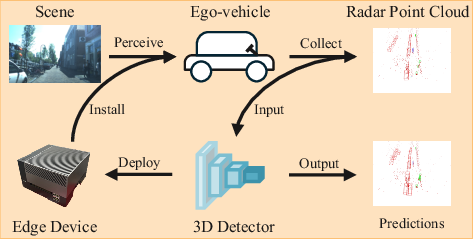}
\caption{Our RadarNeXt enables reliable and efficient real-time 3D detection on edge devices like the Jetson AGX Orin, which has limited computational and energy resources.}
\label{comparisons}
\end{figure}

Several studies use specific algorithms \cite{liu2023smurf}\cite{peng2024mufasa} or modules \cite{musiat2024radarpillars}\cite{guan2024asy} during feature extraction to address the challenges of sparsity and noise in 4D mmWave radar. However, adding these algorithms or modules increases computational complexity and number of parameters, reducing efficiency and hindering their applications on resource-limited edge devices. The sparse data lend themselves to sparse convolutions \cite{yan2018second}\cite{li2023pillarnext} or attention mechanisms \cite{musiat2024radarpillars} for lightweight network design for point-cloud-based 3D detection. However, dense detectors with single-path topologies \cite{lang2019pointpillars} can sometimes outperform sparse detectors \cite{li2023pillarnext} in inference efficiency. Consequently, we reconstruct the single-path backbone of PointPillars with the Re-parameterizable Depthwise Convolution (Rep-DWC) proposed by MobileOne \cite{vasu2023mobileone},  extracting high-quality features while ensuring fast inference. This enables RadarNeXt to operate on edge devices with limited computational resources and energy, providing real-time 3D object detection.

To maintain the efficiency of feature extraction, RadarNeXt avoids applying extra algorithms or modules in the backbone to handle sparsity and noise in 4D mmWave radar point clouds. In several radar-camera fusion-based object detection studies, radar point cloud features are enhanced during multi-modal feature fusion to improve the network performance. Therefore, inspired by \cite{guan2023achelous}\cite{guan2024nanomvg}, we propose the Multi-path Deformable Foreground Enhancement Network (MDFEN) as the neck of RadarNeXt to address the challenges of sparsity and noise in mmWave radar point clouds. MDFEN leverages the adaptive receptive fields and weights of Deformable Convolution v3 (DCNv3) to restore object geometric representations embedded in the feature of sparse and noisy point clouds, improving feature quality. Furthermore, the path aggregation structure \cite{li2022yolov6} fuses the enhanced features with multi-scale information. This increases the depth and robustness of the detector to improve the final performance. Consequently, our RadarNeXt achieves a better trade-off between accuracy and efficiency, providing reliable and real-time 3D object detections.

In conclusion, our work offers three key contributions:
\begin{itemize}
    \item We propose RadarNeXt, a novel and reliable real-time 3D object detection network for 4D mmWave radar point clouds, which achieves 50.48 and 32.30 mAPs on the View-of-Delft (VoD) \cite{palffy2022multi} and TJ4DRadSet (TJ4D) \cite{zheng2022tj4dradset} datasets, with inference speeds of 67.10 FPS on an RTX A4000 GPU and 28.40 FPS on a Jetson AGX Orin, respectively.
    \item To maintain the efficiency of extracting high-quality multi-scale features, we replace the traditional convolution in the PointPillars backbone with the Re-parameterizable Depthwise Convolution (Rep-DWC) proposed by MobileOne. This reduces the number of parameters in our network by 71\% without sacrificing detection accuracy, while improving inference speed by 9\% on the RTX A4000 GPU and 5\% on the Jetson AGX Orin.
    \item To achieve a better trade-off between inference speed and detection accuracy, we introduce the Multi-path Deformable Foreground Enhancement Network (MDFEN) as the neck of RadarNeXt to enhance the quality of features.
\end{itemize}

\section{RELATED WORKS}

\subsection{4D mmWave Radars in 3D Object Detection}

Instead of using traditional networks for automotive radars \cite{zhang2021raddet}\cite{palffy2020cnn}, the similarities between 4D mmWave radar and LiDAR point clouds enable 3D object detection to adopt LiDAR-based detectors \cite{yan2018second}\cite{li2023pillarnext}\cite{lang2019pointpillars}\cite{yin2021center}\cite{wang2020pillar} relying on three techniques: Voxelization, Pillarization, and Multi-View Fusion. However, the extreme sparsity and random noise inherent in mmWave radar prevent LiDAR-based detectors from performing optimally on radar point clouds \cite{venon2022millimeter}. These challenges of sparsity and noise in 4D mmWave radar point clouds have motivated several 4D mmWave radar-based 3D object detection studies to better exploit the advantages of mmWave radar.

Many studies aim to address the sparsity and random noise during the feature extraction stage. \cite{guan2023achelous} and \cite{xu2021rpfa} integrate learnable modules, such as deformable convolutions and attention mechanisms. Leveraging learnable dynamic receptive fields and adaptive weights enhances the network's robustness to sparsity and noise in mmWave radar point clouds. Additionally, some 4D mmWave radar-based detectors \cite{liu2023smurf}\cite{peng2024mufasa} employ specialized algorithms to improve radar point cloud representations. All these methods allow their networks to outperform LiDAR-based 3D object detectors on the 4D mmWave radars.

Furthermore, the sparsity and noise in radar point clouds are addressed during the feature fusion stage. Some \cite{xiong2023lxl}\cite{liu2024mssf} directly use cross-modal information to highlight key representations from the mmWave radar features. In other approaches, the mmWave radar features are first enhanced by attention mechanisms \cite{guan2024nanomvg} or deformable convolutions \cite{guan2024talk2radar} before fusion. These enhanced features improve the final performance of the multi-modal detector, demonstrating the effectiveness of addressing sparsity and noise during the fusion stage. This inspires us to shift the task of overcoming the challenges in mmWave radar to the feature fusion stage.

However, the inclusion of additional modules or modalities slows down the network's inference speed. As a result, most of these approaches fail to meet the real-time requirements for 3D object detection, particularly on edge devices in practical applications. Therefore, in this work, we focus on designing a network that ensures real-time inference capability while addressing the challenges of sparsity and noise in 4D mmWave radar.

\subsection{Efficient 3D Object Detections for Point Clouds}

In traditional 2D object detection studies, the inference speeds of networks can be improved by pruning, knowledge distillation, and quantization \cite{liu2024lightweight}. For sparsely distributed point clouds, a more direct approach is to employ sparse convolutions \cite{yan2018second}\cite{li2023pillarnext} or attention mechanisms \cite{musiat2024radarpillars}, reducing the number of parameters and computations to accelerate the network inference. However, reducing the number of parameters and computational costs does not always lead to improved inference speed of detectors. For example, PillarNeXt \cite{li2023pillarnext}, which employs a multi-path structure and sparse 2D convolutions, is slower than PointPillars \cite{lang2019pointpillars}, which uses a single-path topology and traditional 2D convolutions. This suggests that the network's structural design is just as crucial for inference speed. On the other hand, in image-based 2D object detection, re-parameterizable designs can ensure the reliability and inference efficiency of the network simultaneously \cite{vasu2023mobileone}. Therefore, inspired by PillarNeXt, we adopt the Rep-DWC to reconstruct the PointPillars backbone in our work.

\begin{figure*}[!t]
\centering
\includegraphics[width=7in]{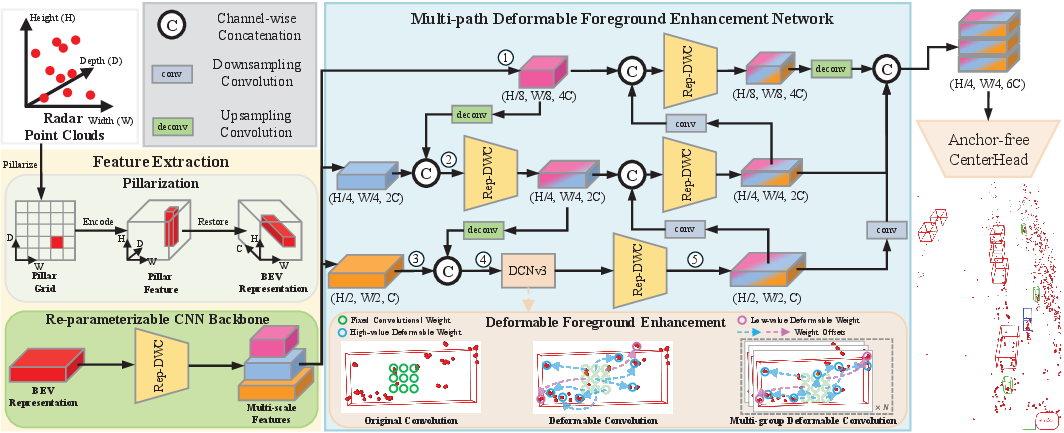}
\caption{The overview of the architecture of our RadarNeXt. \textcircled{1}, \textcircled{2}, \textcircled{3}, \textcircled{4}, and \textcircled{5} are potential positions of DCNv3 in the MDFEN variants for the following ablation studies to validate the current design that DCNv3 on the position \textcircled{4}.}
\label{architecture}
\end{figure*}

\section{METHODOLOGY}

Figure \ref{architecture} illustrates an overview of the RadarNeXt architecture. Pillarization, the faster method than Voxelization and Multi-View Fusion \cite{li2023pillarnext}, encodes raw point features into pseudo-image representations in the BEV perspective. Rep-DWC backbone will extract multi-scale information from the output of Pillarization. During fusing the multi-scale features, our MDFEN will enhance the foreground representations by DCNv3. Finally, PillarNeXt's CenterHead provides the detection results based on the fused feature from MDFEN.

In this section, we will introduce the Rep-DWC backbone and MDFEN neck in RadarNeXt. Meanwhile, CenterHead and the loss functions for training the detector are summarized. 

\subsection{Re-parameterizable Depthwise Convolution (Rep-DWC) Backbone} 
To accelerate the inference of RadarNeXt, particularly on edge devices like the Jetson AGX Orin, we employ MobileOne's Rep-DWC \cite{vasu2023mobileone}, as shown in Figure 3, to reconstruct the single-path backbone of PointPillars.

Comparisons in Figure \ref{comparisons} reveal that inference speed depends on network topology. For instance, despite having fewer parameters and FLOPs, the multi-path PillarNeXt is slower than the single-path PointPillars.

Rep-DWC \cite{vasu2023mobileone} can reduce the number of parameters in the backbone, making the overall network more lightweight and better suited for edge devices with limited computational resources in practical applications. However, reduced parameter counts also compromise the quality of the extracted features. To address this, Rep-DWC employs an over-parameterization design during training. As shown in Figure 3, the down-sampling layers during training consist of two branches, while the sampling layers contain three. These branches extract representations at various scales, supplementing the feature information and ensuring the quality of the final multi-scale features.

\begin{figure}[!t]
\centering
\includegraphics[width=2.8in]{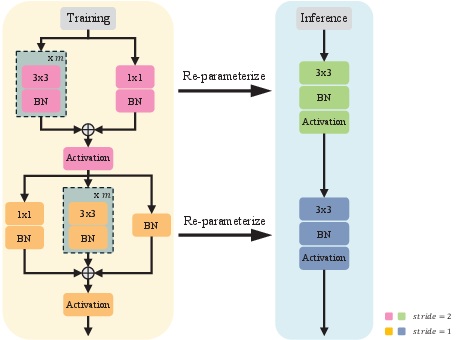}
\caption{The architecture of Rep-DWC proposed by MobileOne. Pink and green are down-sampling layers with “$stride=2$”. Orange and Blue are normal layers with “$stride=1$”.}
\label{rep-dwc}
\end{figure}

During inference, Rep-DWC is transformed into a single-path topology through re-parameterization \cite{vasu2023mobileone}, as shown on the right of Figure 3. Re-parameterization aggregates the convolutions and batch normalizations of each branch in the training phase into a single convolution and batch normalization. This transformation maintains the feature extraction ability, ensuring feature quality during inference despite the change in network topology.

Therefore, the Rep-DWC backbone accelerates inference while maintaining the accuracy of RadarNeXt, enabling reliable and real-time 3D detections for practical applications.

\subsection{Multi-path Deformable Foreground Enhancement Network (MDFEN)}

To address the challenges of sparsity and random noise in millimeter-wave radar point clouds, we propose a Multi-path Deformable Foreground Enhancement Network (MDFEN). As shown in Figure \ref{architecture}, MDFEN combines the DCNv3 \cite{wang2023internimage} and Path Aggregation Network (PAN) \cite{li2022yolov6} to adaptively select and aggregate features using dynamic receptive fields and weights, enhancing foreground object representations.

We apply DCNv3 only to the largest-scale feature map before fusing it with a single Rep-DWC layer, balancing spatial information enhancement with computational costs. DCNv3 \cite{wang2023internimage} aggregates results from $N$ adaptive receptive fields and weights, restoring spatial information in the concatenated feature map before completing top-down multi-scale fusion to enhance object representations.

\begin{align}
    y(z_0) = \sum\limits^{N}_{n=1}\sum\limits^{K}_{k=1}w_nm_{nk}x_n(z_0+z_k+\Delta p_{nk})
\label{eq:dcn}
\end{align}

where $N$ is the number of groups for aggregations, and $K$ is the number of sampling points in the $n$-th group. $y(z_0)$ is the output of DCNv3 whose kernel locates at $z_0$, while the $x_n$ as the input of the $n$-th group of DCNv3. $z_k$ denotes the original weight location in the DCNv3 kernel, offset by the learnable parameter $\Delta p_{nk}$. The corresponding weight value at location $z_k$ is $w_n$, which is modulated by another learnable parameter $m_{nk}$ to adjust the impact of the captured feature.

Subsequently, the Rep-DWC layer encodes the enhanced foreground representations into a feature map of size (H/2, W/2, C), where H, W, and C denote the height, width, and channels of the pseudo-image feature map from Pillarization. Finally, during PAN’s bottom-up fusion phase, the features from DCNv3 are fused with multi-scale feature maps, integrating enhanced spatial information with spatial and multi-scale semantic representations in the high-dimensional feature maps.

In Figure \ref{architecture}, we employ a multi-path topology in the training phase to offset performance degradation due to DWC's low parameter counts \cite{vasu2023mobileone}. On the other hand, to further improve the quality of multi-scale fusion features, we incorporate an additional Feature Pyramid Network (FPN) \cite{lang2019pointpillars} after PAN to deepen our RadarNeXt.

\subsection{Anchor-free CenterHead and Loss Functions}

We adopt the anchor-free CenterHead from PillarNeXt for detection results. This head utilizes Focal Loss $L_{\text{FL}}$ \cite{ross2017focal} and L1 Loss $L_{\text{L1}}$ \cite{qi2020mean} to optimize object classification and the regression of corresponding 3D bounding boxes. Furthermore, PillarNeXt introduces IoU Loss $L_{\text{IoU}}$ and dIoU Loss $L_{\text{dIoU}}$ to aid in the regression of 3D bounding boxes, improving detection accuracy.

\begin{align}
    L_{\text{IoU}} = \sum |IoU_p-IoU_r|
\label{eq:iou}
\end{align}

\begin{align}
    L_{\text{dIoU}} = 1-dIoU
\label{eq:diou}
\end{align}

where $IoU_p$ is the IoU value predicted by the network. $IoU_p$ and $dIoU$ are metrics for the overlap between predicted 3D bounding boxes $p_{\text{box}}$ and ground-truth boxes $g_{\text{box}}$, where dIoU also measures the Euclidean distance between the centers and the diagonal differences of the boxes. Inspired by CenterFormer \cite{zhou2022centerformer}, we apply the corner classification loss $L_{\text{mse}}$ based on the MSE Loss during training.

Ultimately, the loss function $L$ of RadarNeXt combines the five aforementioned losses and adjusts their contributions to network learning using a set of weights, denoted as $\Lambda=\{\lambda_{\text{FL}},\ \lambda_{\text{L1}},\ \lambda_{\text{IoU}},\ \lambda_{\text{dIoU}},\ \lambda_{\text{mse}}\}$.

\begin{align}
    L= \lambda_{\text{FL}}L_{\text{FL}}+\lambda_{\text{L1}}L_{\text{L1}}+\lambda_{\text{IoU}}L_{\text{IoU}}+\lambda_{\text{dIoU}}L_{\text{dIoU}}+\lambda_{\text{mse}}L_{\text{mse}}
\label{eq:loss}
\end{align}

\section{EXPERIMENTS AND ANALYSIS}

\subsection{Datasets and Evaluation Metrics}

To demonstrate that our RadarNeXt achieves a better trade-off between reliability and efficiency in 3D object detection with millimeter-wave radar, we compare it with existing 3D point-cloud-based detectors by experiments on two public 4D millimeter-wave imaging radar datasets: VoD \cite{palffy2022multi} and TJ4D \cite{zheng2022tj4dradset}.

VoD provides millimeter-wave radar point clouds in three density levels: single-scan, three-scan, and five-scan. For training and validation, we employ the five-scan dataset, which contains over 8,600 high-density point clouds with 3D bounding box annotations for more than 26,000 cars, 26,000 pedestrians, and 10,000 cyclists. All raw radar point features from VoD are used as inputs $I_{\text{VoD}}$, including 3D coordinates $(x,\ y,\ z)$, Radar Cross Section $RCS$, relative radial velocity $v_r$, absolute radial velocity $v_{ra}$, and timestamp $t$.

\begin{align}
    I_{\text{VoD}} = \left[x,\ y,\ z,\ RCS,\ v_r,\ v_ra,\ t \right]
\label{eq:in_vod}
\end{align}

On the other hand, the TJ4D dataset offers a richer variety of driving scenes and object classes than VoD. It includes over 7,500 frames of radar point clouds with 3D bounding box annotations for more than 16,000 cars, 4,200 pedestrians, 7,300 cyclists, and 5,300 trucks. To generate the 64-channel pillar features, similar to VoD preprocessing, we select six raw TJ4D point features $I_{\text{TJ4D}}$: 3D coordinates $(x,\ y,\ z)$, relative radial velocity $v_r$, detection range $R$, and signal-to-noise ratio $SNR$.

\begin{align}
    I_{\text{TJ4D}} = \left[x,\ y,\ z,\ v_r,\ R,\ SNR \right]
\label{eq:in_tj4d}
\end{align}

We apply the official VoD evaluation script to measure object detection accuracy through Average Precision (AP) and assess overall network reliability by mAP. Moreover, this script can also calculate the mAP to evaluate the overall network reliability. For VoD, the IoU thresholds for calculating AP in VoD are set to 0.5 for cars and 0.25 for both pedestrians and cyclists. For TJ4D, IoU thresholds are set to 0.5 for cars and trucks, and 0.25 for pedestrians and cyclists. The 3D detection performance of the networks is represented by AP for individual classes and mAP for overall accuracy. Notably, the evaluations of the detector's inference speed measured in FPS are on both a server (on
A4000) and an edge device (on Orin).

\subsection{Implementation Details}
\begin{table*}[t!]
\centering
\caption{Comparison on the validation set of VoD dataset and on the test set of TJ4D dataset. For fair comparisons, all the networks are trained and evaluated with 5-scans radar points of VoD and single-scan radar points of TJ4D.}
\vspace{-5pt}
\label{results}
\begin{tabular}{c|c|cccc|ccccc|cc}
\hline
\multirow{2}{*}{\textbf{Networks}} & \multirow{2}{*}{\textbf{Params}} & \multicolumn{4}{c|}{\textbf{VoD} (5-scans)} & \multicolumn{5}{c|}{\textbf{TJ4D}} & \multicolumn{2}{c}{\textbf{FPS}} \\
\cline{3-6} \cline{7-11} \cline{12-13}
 & & Car & Ped & Cyc & $\text{mAP}_{\text{3D}}$ & Car & Ped & Cyc & Truck & $\text{mAP}_{\text{3D}}$ & on A4000 & on Orin\\
\hline
SECOND \cite{yan2018second} & 4.320M & 35.98 & 28.91 & 53.29 & 39.39 & \textit{26.53} & 24.20 & \textit{56.92} & 6.37 & 28.51 & 43.63 & 22.33 \\
PointPillars \cite{lang2019pointpillars} & 4.237M & \textit{41.11} & 39.13 & 65.91 & 48.72 & \textbf{29.03} & 23.53 & 51.27 & 16.54 & 30.09 & \textit{67.40} & 27.17 \\
CenterPoint \cite{yin2021center} & 5.050M & 36.16 & 37.03 & 64.30 & 45.83 & 17.34 & 25.89 & 51.95 & 16.41 & 27.90 & 49.93 & 23.53 \\
PillarNeXt \cite{li2023pillarnext} & 2.083M & 32.04 & 26.85 & 61.06 & 39.98 & 12.98 & 11.34 & 35.32 & 13.42 & 18.26 & 37.27 & 18.33 \\
\hline
SMURF \cite{liu2023smurf} & 17.547M & \underline{42.31} & 39.09 & 71.50 & \textbf{50.97} & \underline{28.47} & \textit{26.22} & 54.61 & \underline{22.64} & \textbf{32.99} & 24.70 & 11.60 \\
LXL-R \cite{xiong2023lxl} & 5.009M & 32.75 & \textit{39.65} & 68.13 & 46.84 & - & - & - & - & 30.79 & 48.67 & 25.33 \\
RadarPillars \cite{musiat2024radarpillars} & - & 41.10 & 38.60 & \underline{72.60} & \underline{50.70} & - & - & - & - & - & - & - \\
MSFF-V-R \cite{liu2024mssf} & - & 38.28 & \textbf{42.93} & 69.96 & 50.39 & 12.34 & \textbf{31.73} & 53.16 & 9.15 & 26.60 & - & - \\
MUFASA \cite{peng2024mufasa} & - & \textbf{43.10} & 38.97 & 68.65 & 50.24 & 23.56 & 23.70 & 48.39 & \textbf{25.25} & 30.23 & - & - \\
\hline
\textbf{RadarNeXt (FPN)} & \textbf{0.899M} & 37.96 & 38.28 & 67.69 & 47.98 & 18.64 & \underline{28.67} & \textbf{63.88} & 16.17 & \textit{31.84} & \textbf{83.57} & \textbf{33.07} \\
\textbf{RadarNeXt (PAN)} & \underline{1.531M} & 35.47 & 36.1 & \textbf{72.89} & 48.15 & 20.49 & 26.10 & 54.90 & 18.02 & 29.88 & \underline{70.97} & \underline{31.43}  \\
\textbf{RadarNeXt (MDFEN)} & \textit{1.580M} & 37.44 & \underline{41.83} & \textit{72.16} & \textit{50.48} & 26.24 & 24.55 & \underline{59.78} & \textit{18.64} & \underline{32.30} & 67.10 & \textit{28.40}  \\
\hline
\end{tabular}
\captionsetup{font=small}
\caption*{\textbf{Bold} is the best, and \underline{underline} is the second best, and \textit{italic} is the third best. Networks are divided into three groups: the first contains those designed for LiDAR, and the second presents the networks with specific designs for mmWave radar. Our RadarNeXts with various necks (FPN, PAN, and MDFEN) are in the last group.}
\end{table*}

\textit{1) Hyper-parameter Setting}: To restore VoD 4D mmWave radar point clouds into pseudo-image representations, we set each pillar within the size of 0.16 meters in length, 0.16 meters in width, and 5 meters in height, deriving a BEV feature map of size $320 \times 320 \times 64$ from Pillarization. For TJ4D, we set the pillar size of 0.16 meters in length, 0.16 meters in width, and 6 meters in height, encoding raw point clouds into a pseudo-image feature of size $432 \times 432 \times 64$. Additionally, voxel sizes for SECOND and SMURF are set to $\text{0.05m (length)
}\times\text{0.05m (width)}\times\text{0.1m (height)}$ and $\text{0.16m (length)
}\times\text{0.16m (width)}\times\text{0.2m (height)}$, respectively. For TJ4D, voxel sizes are adjusted to $\text{0.04m (length)
}\times\text{0.04m (width)}\times\text{0.12m (height)}$ for SECOND and $\text{0.16m (length)
}\times\text{0.16m (width)}\times\text{0.24m (height)}$ for SMURF.

The details of our RadarNeXt are as follows: The number of channels of the Pillarization outputs is set to $64$, denoted by $C=64$. Similar to PointPillars, the Rep-DWC backbone comprises three stages to extract feature maps at different scales. The first stage is composed of three Rep-DWC layers, and the rest has five layers, including one layer for down-sampling the feature maps to obtain multi-scale representations in each stage. in MDFEN, a single-layer Rep-DWC module is employed for channel-wise feature fusion. All the Rep-DWC layers use $m=1$.

Finally, all convolutional kernel sizes, including DCNv3, are set to $3 \times 3$, except for the $1 \times 1$ Depthwise and Pointwise Convolutions in Figure \ref{rep-dwc} and the Upsampling Convolution in Figure \ref{architecture}. The Upsampling Convolution uses a kernel size of $2$ and stride of $2$ to satisfy upsampling requirements. The number of groups $N$ in DCNv3 is set to 4, enabling the aggregation of multiple foreground enhancements without significantly increasing the network's computational cost.

\textit{2) Training Setting}: Since both VoD and TJ4D datasets follow the KiTTi format, we train RadarNeXt using the MMDetection3D toolbox \cite{mmdet3d2020} on an Nvidia RTX A4000 GPU for 80 epochs with a batch size of 8. The network weights are saved every five epochs to identify the best detection performance, and the weights achieving this performance are used as the final training results. The learning rate strategy is step decay, with the initial learning rate set to 3e-3. The AdamW is employed as the optimizer. Moreover, to enhance the robustness of detectors in various driving scenes, we use the global scale and random flip along with the x-axis as data augmentation strategies.

\subsection{Quantitative Results}

\begin{figure*}[!t]
\centering
\includegraphics[width=7in]{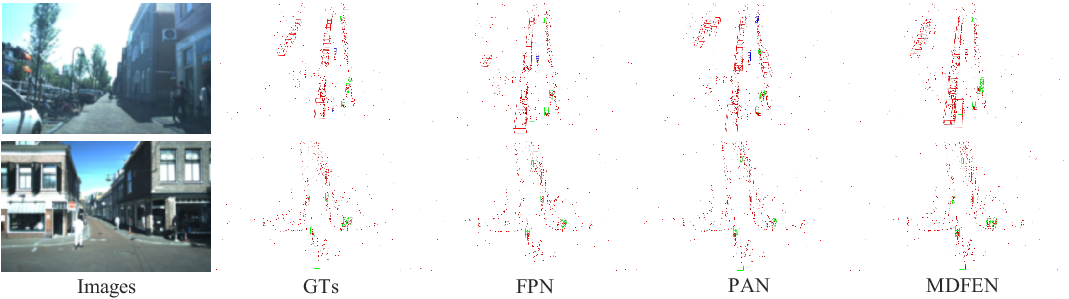}
\caption{Qualitative results on the VoD dataset: Images are the image of driving scenes. GTs are the ground truths. The red dots within figures are radar points. Red cuboids represent the Car. Blue cuboids represent the Cyclist. Green cuboids represent the Pedestrian. FPN shows the results of RadarNeXt (FPN). PAN shows the results of RadarNeXt (PAN). MDFEN shows the results of RadarNeXt (MDFEN).}
\label{qua_result_vod}
\end{figure*}

\begin{figure*}[!t]
\centering
\includegraphics[width=7in]{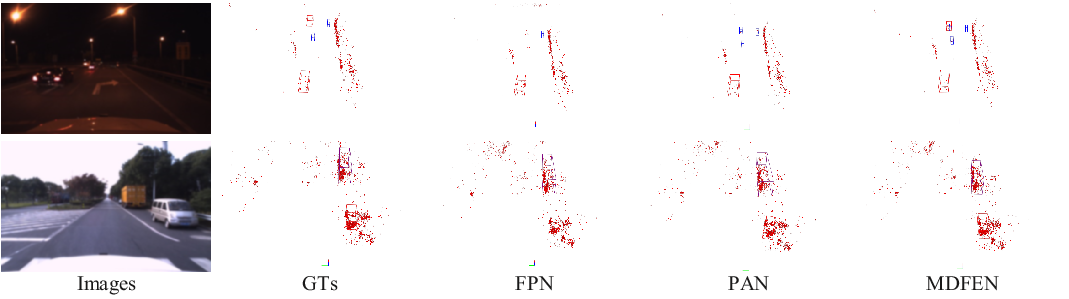}
\caption{Qualitative results on the TJ4D dataset: Images are the image of driving scenes. GTs are the ground truths. The red dots within figures are radar points. Red cuboids represent the Car. Blue cuboids represent the Cyclist. Green cuboids represent the Pedestrian. Purple cuboids represent the Truck. FPN shows the results of RadarNeXt (FPN). PAN shows the results of RadarNeXt (PAN). MDFEN shows the results of RadarNeXt (MDFEN).
}
\label{qua_result_tj4d}
\end{figure*}

In Table \ref{results}, we compare RadarNeXt with LiDAR-based and mmWave radar-based 3D object detectors on the validation set of VoD and the test set of TJ4D. Our detector, represented by \textbf{RadarNeXt (MDFEN)}, achieves 50.48 and 32.30 mAP, respectively, with only 1.580M parameters. This performance surpasses all LiDAR-based detectors in the first group of Table \ref{results}. Moreover, in comparison to the mmWave radar-based 3D detectors in the second group, RadarNeXt ranks third on VoD and second on TJ4D. This demonstrates the effectiveness of our strategy for enhancing the foreground representations within mmWave radar features during multi-scale feature fusion.

Additionally, the inference speed of \textbf{RadarNeXt (MDFEN)} reaches 67.10 FPS on the RTX A4000 GPU and 28.40 FPS on the Jetson AGX Orin, as shown in Table 1. This inference efficiency ensures the real-time capability of our network, particularly for edge devices like the Orin. Furthermore, to demonstrate the contribution of the Rep-DWC backbone to network efficiency and real-time inference, we implement two RadarNeXt variants using FPN and PAN as necks, denoted as \textbf{RadarNeXt (FPN)} and \textbf{RadarNeXt (PAN)}. Among previous 3D detectors, PointPillars, using FPN as the neck, achieves the fastest inference speed of 67.40 FPS on the A4000. The efficiency of \textbf{RadarNeXt (FPN)} with the Rep-DWC backbone is 24\% faster than PointPillars, reaching 83.57 FPS on the A4000 and maintaining 33.07 FPS on the Orin. Notably, \textbf{RadarNeXt (FPN)}, the fastest object detector in the table, achieves 31.84 mAP on the TJ4D, ranked third among all the networks. This indicates that the Rep-DWC backbone accelerates the inference while ensuring the reliability of RadarNeXt.

However, \textbf{RadarNeXt (FPN)} achieves only 47.98 mAP on the VoD validation set, and the deeper \textbf{RadarNeXt (PAN)} with PAN achieves only 48.15 mAP on the VoD. The accuracy of both RadarNeXt variants is lower than that of PointPillars in the first group of Table \ref{results}. \textbf{RadarNeXt (MDFEN)} utilizes MDFEN for foreground enhancement, achieving the third-best accuracy on VoD and the second-best on TJ4D. Meanwhile, its inference speed on the A4000 GPU is comparable to that of PointPillars. Additionally, \textbf{RadarNeXt (MDFEN)} is more efficient than PointPillars on the Orin, ranking thrid-fastest behind the two variants with more than 30 FPS. These results validate the reliability of our MDFEN in addressing the challenges of sparsity and noise in mmWave radar point clouds. MDFEN contributes to achieving a better balance between accuracy and efficiency.

Figures \ref{qua_result_vod} and \ref{qua_result_tj4d} show the detection results on the VoD and TJ4D derived from three networks: \textbf{RadarNeXt (FPN)}, \textbf{RadarNeXt (PAN)}, and \textbf{RadarNeXt (MDFEN)}. By comparing these results, we highlight the advantages of MDFEN in 3D object detection for 4D mmWave radar point clouds. As shown in Figure \ref{qua_result_vod}, RadarNeXt with MDFEN detects distant targets (Cars and Pedestrians) more accurately than the other two variants on the VoD dataset. The results in two driving scenes validate that MDFEN can overcome challenges posed by sparse point distributions through dynamic receptive fields of DCNv3 layers,  enhancing object geometric information. Moreover, in addition to demonstrating the advantages of MDFEN for long-distance object detection in the TJ4D scene at the top of Figure \ref{qua_result_tj4d}, the results for the scene at the bottom show that MDFEN also overcomes noise influence. \textbf{RadarNeXt (FPN)} and \textbf{RadarNeXt (PAN)} variants miss the detection or provide an inaccurate prediction for the car nearby. In contrast, \textbf{RadarNeXt (MDFEN)} detects it accurately. This improvement is primarily due to the learnable weights from DCNv3, filtering out noises effectively.

\subsection{Ablation Studies}

Ablation studies will further validate the effectiveness of our design using the VoD validation set. Additionally, we will evaluate the networks in each study on the RTX A4000 GPU and Jetson AGX Orin to fully assess the contribution of the Rep-DWC backbone and MDFEN neck.

Table \ref{ablation_backbone} compares the Rep-DWC backbone with the PointPillars backbone (denoted as Dense) on the VoD validation set, using the three RadarNeXt variants: \textbf{RadarNeXt (FPN)}, \textbf{RadarNeXt (PAN)}, and \textbf{RadarNeXt (MDFEN)}. Replacing Dense with \textbf{RadarNeXt (FPN)} increases the parameter size to 4.726M. These additional parameters improve model accuracy to 48.89 mAP, but the inference speed drops to 72.33 FPS on the A4000 and 30.83 FPS on Orin. The performance of the Dense variant corresponding to \textbf{RadarNeXt (PAN)} drops to 47.72 mAP compared to the \textbf{RadarNeXt (FPN)} dense variant. This result demonstrates that increasing parameters and depth does not effectively improve detector reliability. Consequently, the accuracy of the Dense variant corresponding to \textbf{RadarNeXt (MDFEN)} is only 48.23 mAP, lower than the 50.48 mAP achieved by our RadarNeXt on VoD. Moreover, the additional parameters decrease the inference efficiency of networks. Therefore, the Rep-DWC backbone better accelerates network execution while maintaining detection reliability in conjunction with MDFEN.

\begin{table}[t!]
\centering
\caption{Comparison Rep-DWC backbone with PointPillars backbone on the VoD validation set. All the networks are evaluated on 5-scans radar points for fair comparison.}
\vspace{-5pt}
\label{ablation_backbone}
\begin{tabular}{c|c|c|cc}
\hline
\multirow{2}{*}{\textbf{Networks}} & \multirow{2}{*}{\textbf{Params}} & \textbf{VoD} & \multicolumn{2}{c}{\textbf{FPS}} \\
\cline{3-5}
 & & $\text{mAP}_{\text{3D}}$ & on A4000 & on Orin  \\
\hline
Dense + FPN & 4.726M & 48.89 & 72.33 & 30.83 \\
\textbf{RadarNeXt (FPN)} & 0.899M & 47.98 & 83.57 & 33.07 \\
\hline
Dense + PAN & 5.337M & 47.72 & 69.07 & 29.63 \\
\textbf{RadarNeXt (PAN)} & 1.531M & 48.15 & 70.97 & 31.43 \\
\hline
Dense + MDFEN & 5.385M & 48.23 & 61.43 & 27.13 \\
\textbf{RadarNeXt (MDFEN)} & 1.580M & 50.48 & 67.10 & 28.40 \\
\hline
\end{tabular}
\captionsetup{font=small}
\caption*{The variant with PointPillars backbone is in the same group with its corresponding RadarNeXt, represented by "Dense + FPN," "Dense + PAN," and "Dense + MDFEN," respectively.}
\end{table}

\begin{table}[t!]
\centering
\caption{Ablation Study of MDFEN variants with various positions of DCNv3. All the variants are evaluated on the 5-scans radar points for fair comparisons.}
\vspace{-5pt}
\label{ablation_neck}
\begin{tabular}{c|c|c|cc}
\hline
\multirow{2}{*}{\textbf{DCNv3 Positions}} & \multirow{2}{*}{\textbf{Params}} & \textbf{VoD} & \multicolumn{2}{c}{\textbf{FPS}} \\
\cline{3-5}
 & & $\text{mAP}_{\text{3D}}$ & on A4000 & on Orin  \\
\hline
\textcircled{1} & 1.694M & 48.09 & 69.30 & 29.73 \\
\textcircled{3} & 1.630M & 48.35 & 68.33 & 30.13 \\
\textcircled{5} & 1.547M & 49.13 & 68.90 & 30.60 \\
\hline
\textcircled{2} & 1.694M & 47.45 & 68.30 & 28.80 \\
\textcircled{4} & 1.580M & 50.48 & 67.10 & 28.40 \\
\hline
\textcircled{1}\textcircled{2}\textcircled{4} & 1.905M & 46.96 & 59.87 & 25.50 \\
\hline
\end{tabular}
\captionsetup{font=small}
\caption*{Figure \ref{architecture} displays the DCNv3 position in each variant indicated by \textcircled{1}, \textcircled{2}, \textcircled{3}, \textcircled{4}, and \textcircled{5}. DCNv3 on \textcircled{4} is our final design for MDFEN, and \textcircled{1}\textcircled{2}\textcircled{4} denotes the variant with three DCNv3 layers on corresponding positions.}
\end{table}

On the other hand, we experiment with placing DCNv3 layers in different positions to validate the effectiveness of the current MDFEN design, as shown in Table \ref{ablation_neck}. Firstly, we use DCNv3 solely to process normal feature maps, such as the outputs of the Rep-DWC backbone with 64 and 256 channels and the fusion layer output with 64 channels. These configurations improve RadarNeXt's inference speed, particularly on Orin, reaching a maximum of 30.60 FPS. However, compared to the MDFEN design in \textbf{RadarNeXt (MDFEN)}, the network's detection accuracy drops to a maximum of 49.13 mAP. Moreover, we try to place the DCNv3 layers at the \textcircled{2} position in Figure \ref{architecture}. This attempt slightly improves speed to 68.30 FPS on the A4000 and 28.80 FPS on Orin, but significantly reduces the detection accuracy of RadarNeXt. Finally, we evaluate using DCNv3 to process all feature maps before inputting them into the fusion layers. This variant is denoted as \textcircled{1}\textcircled{2}\textcircled{4}. As a result, the additional DCNv3 layers reduce inference speed to 59.87 FPS on the A4000 and 25.50 FPS on Orin. These sacrifices did not improve accuracy, 46.96 mAP on the VoD validation set. Therefore, these results demonstrate that our current MDFEN design, with the DCNv3 positioned at \textcircled{4}, better balances accuracy and efficiency. Our design makes RadarNeXt reliable enough for practical real-time 3D object detection applications using 4D mmWave radar point clouds.

\section{CONCLUSIONS}

In this paper, we propose RadarNeXt, a real-time and reliable 3D object detector tailored for 4D mmWave imaging radar. To accelerate inference, we use the Rep-DWC backbone and defer the processing of sparsity and noise in 4D mmWave radar point clouds to the feature fusion stage. Subsequently, we propose MDFEN as the neck to achieve a better trade-off between accuracy and efficiency, enabling reliable and real-time 3D object detection for practical applications. Finally, RadarNeXt achieves 50.48 mAP on the VoD and 32.30 mAP on the TJ4D, with 67.10 FPS on the RTX A4000 GPU and 28.40 FPS on the Jetson AGX Orin. This accuracy is comparable to that of SMURF, while the speed on Orin is 5\% faster than PointPillars. In the future, we plan to introduce efficient attention mechanisms in the feature extraction stage to selectively encode point features, aiming to improve detection accuracy for stationary objects.


\section*{ACKNOWLEDGMENT}

This work is partially supported by the XJTLU AI University Research Centre and Jiangsu Province Engineering Research Centre of Data Science and Cognitive Computation at XJTLU. Also, it is partially funded by the Suzhou Municipal Key Laboratory for Intelligent Virtual Engineering (SZS2022004) as well as funding: XJTLU-REF-21-01-002, XJTLU-RDF-22-01-062, and XJTLU Key Program Special Fund (KSF-A-17).

This work received financial support from Jiangsu Industrial Technology Research
Institute (JITRI) and Wuxi National Hi-Tech District (WND).

\bibliographystyle{IEEEtran} 
\bibliography{ref}

\end{document}